%% file: root.tex
\documentclass[letterpaper, 10 pt, conference]{ieeeconf}  %

\input{custom_commands.tex}
\IEEEoverridecommandlockouts                              %

\usepackage{color, colortbl}
\usepackage[dvipsnames]{xcolor}
\usepackage{graphicx}
\usepackage{booktabs}
\usepackage{mathtools}
\usepackage{float}
\usepackage{amsfonts}
\usepackage{algorithm}
\usepackage{algpseudocode}
\usepackage{hyperref}
\usepackage{sidecap}
\usepackage{tabularray}

\usepackage[belowskip=0pt,aboveskip=5pt,font=small]{caption}
\usepackage{lipsum}
\usepackage{amsmath}
\usepackage{url}
\hypersetup{
    colorlinks=true,
    linkcolor=blue,
    filecolor=magenta,      
    urlcolor=cyan,
    pdftitle={Overleaf Example},
    pdfpagemode=FullScreen,
    }

\title{\LARGE \bf
EgoMimic: Scaling Imitation Learning via Egocentric Video
}

\author{Simar Kareer$^1$, Dhruv Patel$^{1*}$, Ryan Punamiya$^{1*}$, Pranay Mathur$^{1*}$, Shuo Cheng$^1$ \\ Chen Wang$^2$, Judy Hoffman\textsuperscript{1\textdagger}, Danfei Xu\textsuperscript{1\textdagger} %
\thanks{\textsuperscript{1} Georgia Institute of Technology, GA, USA.}
\thanks{\textsuperscript{2} Stanford University, CA, USA.}%
\thanks{\textsuperscript{*}Equal contribution. }
\thanks{\textsuperscript{\textdagger}Equal advising. }
}

\begin{document}

\twocolumn[{%
  \renewcommand\twocolumn[1][]{#1}%
  \maketitle
  \thispagestyle{empty}
  \pagestyle{empty}
  \vspace{-0.5cm}
  \includegraphics[width=\textwidth]{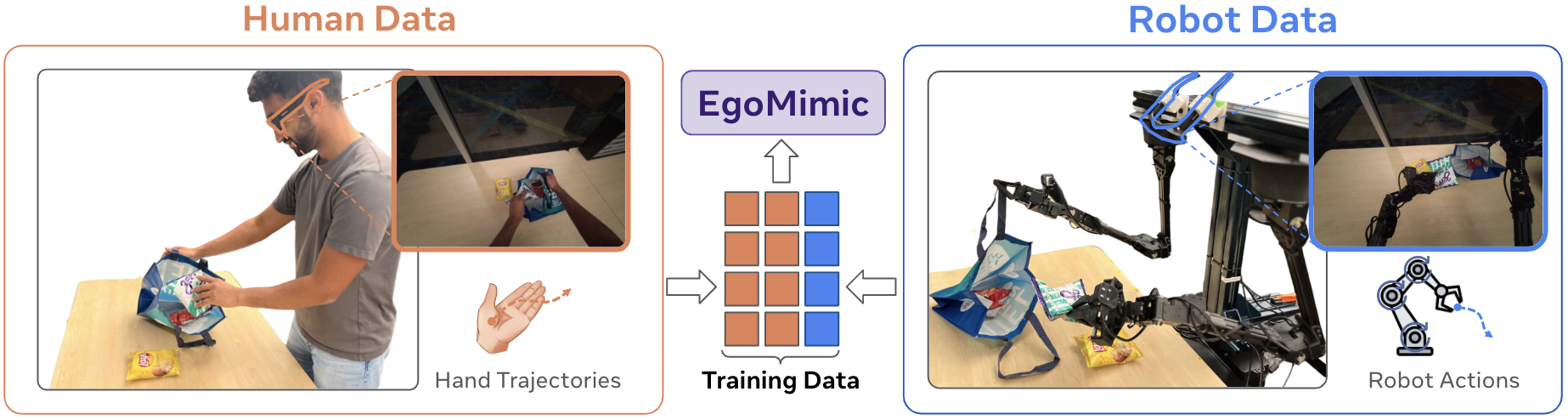}
  \vspace{-0.45cm}
  \captionof{figure}{
\method unlocks \emph{human embodiment data}---egocentric videos paired with 3D hand tracks---as a new scalable data source for imitation learning. We can capture this data anywhere, without a robot, by wearing a pair of Project Aria glasses while performing manipulation tasks with our own hands. \method bridges kinematic, distributional, and appearance differences between human embodiment data (left) and traditional robot teleoperation data (right) to learn a unified policy. We find that human embodiment data boosts task performance by 34-228\% over using robot data alone, and enables generalization to new objects or even scenes.}
  \label{fig:teaser}
  \vspace{0.5cm}
}]

\let\svthefootnote\thefootnote
\let\thefootnote\relax\footnote{SK, DP, RP, PM, SC, JH, DX are with the Georgia Institute of Technology and CW is with Stanford University.  Email Correspondence: {\tt\small skareer@gatech.edu}}
\let\thefootnote\relax\footnote{\llap{\textsuperscript{*}}Denotes equal contribution. \llap{\textsuperscript{\textdagger}}Denotes equal advising.}
\addtocounter{footnote}{-1}\let\thefootnote\svthefootnote\

\vspace{-0.35cm}
\begin{abstract}
The scale and diversity of demonstration data required for imitation learning is a significant challenge. We present \method, a full-stack framework which scales manipulation via human embodiment data, specifically egocentric human videos paired with 3D hand tracking.
\method achieves this through: (1) 
a system to capture human embodiment data using the ergonomic Project Aria glasses, (2) a low-cost bimanual manipulator that minimizes the kinematic gap to human data, (3) cross-domain data alignment techniques, and (4) an imitation learning architecture that co-trains on human and robot data. Compared to prior works that only extract high-level intent from human videos, our approach treats human and robot data equally as embodied demonstration data and learns a unified policy from both data sources. \method achieves significant improvement on a diverse set of long-horizon, single-arm and bimanual manipulation tasks %
over state-of-the-art imitation learning methods and enables generalization to entirely new scenes.  Finally, we show a favorable scaling trend for \method, where adding 1 hour of additional hand data is significantly more valuable than 1 hour of additional robot data. Videos and additional information can be found at \url{https://egomimic.github.io/}

\end{abstract}

\nopagebreak
\input{sections/intro}

\input{sections/relatedWorks}

\input{sections/method}

\input{sections/experimentalSetup}

\input{tables/success_rate}

\input{tables/ablations}
\input{sections/results}

\input{sections/conclusion}

\bibliographystyle{IEEEtran}
\bibliography{main}
\clearpage

\input{sections/appendix}

\end{document}

%% file: custom_commands.tex
\newcommand{\tline}[1]{\specialrule{1px}{0.2em}{0.2em}}
\newcommand{\mline}[1]{\specialrule{0.5px}{0.2em}{0.2em}}

\usepackage{xspace}

\newcommand{\simar}[1]{{\textcolor[rgb]{0.0, 0.0,0.0}{#1}}}

\newcommand{\method}{EgoMimic\xspace}
\newcommand{\price}{\$1,000\xspace}
\newcommand{\aria}{Aria glasses\xspace}

%% file: sections/intro.tex
\section{Introduction}\label{intro}

End-to-end imitation learning has shown remarkable performance in learning complex manipulation tasks, but it remains brittle when facing new scenarios and tasks. Drawing on the recent success of Computer Vision and Natural Language Processing, we hypothesize that for learned policies to achieve broad generalization, we must dramatically scale up the training data size. While these adjacent domains benefit from Internet-sourced data, robotics lacks such an equivalent.

To scale up data for robotics, there have been recent advances in data collection systems.  For example, ALOHA~\cite{zhao2023act,aloha2} and GELLO~\cite{wu2024gellogenerallowcostintuitive} are intuitive leader-follower controls for collecting teleoperated data. Other works have opted to develop hand-held grippers to collect data without a robot~\cite{chi2024universalmanipulationinterfaceinthewild}. 
Despite these advances, data collected via these systems still require specialized hardware and active effort in providing demonstrations. %
We hypothesize that a key step for achieving Internet-scale robot data is \emph{passive data collection}. Just as the Internet was not built for curating data to train large vision and language models, an ideal robot data system should allow users to generate sensorimotor behavior data without intending to do so.  %

Human videos, especially those captured from an egocentric perspective, present an ideal source of data for passive data scalability. {This data aligns closely with robot data, as it provides an
egocentric camera for vision, 3D hand tracking for actions, and onboard SLAM for localization.}
The advent of consumer-grade devices capable of capturing such data, including Extended Reality (XR) devices and camera-equipped ``smart glasses'', opens up unprecedented opportunities for passive data collection at scale.  While recent works have begun to leverage human video data, their approaches are limited to extracting high-level intent information from videos to build planners that guide low-level conditional policies~\cite{wang2023mimicplay,bharadhwaj2023hopman}. As a result, these systems remain constrained by the performance of low-level policies, which are typically trained solely on teleoperation data. 

We argue that to truly scale robot performance with human data, we should not consider human videos as an auxiliary data source that requires separate handling. Instead, we should exploit the inherent similarities between egocentric human data and robot data to treat them as equal parts in a continuous spectrum of embodied data sources. Learning seamlessly from both data sources will require full-stack innovation, from data collection systems that unify data from both sources to imitation learning architectures that can enable such cross-embodied policy learning.

\simar{To this end, our work treats human data as a \emph{first-class data source} for robot manipulation.  We believe our system is a key step towards using passive data from wearable smart glasses to train manipulation policies.  We present \method (Fig.~\ref{fig:teaser}), a framework to collect data and co-train manipulation policies from both human egocentric videos and teleoperated robot data consisting of:}

\textbf{(i)} A system to collect human data built on Project Aria glasses~\cite{engel2023projectarianewtool} that capture egocentric video, 3D hand tracking, and device SLAM. This rich information allows us to transform human egocentric data into a format compatible with robot imitation learning. %

\textbf{(ii)} A capable yet low-cost bimanual robot that minimizes the kinematic and camera-to-camera gap to human embodiment data.  %
In particular, we minimize the camera-to-camera device gap (FOV, dynamic ranges, etc) between human and robot data by using Project Aria glasses as the main robot sensor.

\textbf{(iii)} To mitigate differences in data distributions, we normalize and align action distributions between human and robots.  Further, we minimize the appearance gap between human arm and robot manipulator via visual masking.

\textbf{(iv)} A unified imitation learning architecture that co-trains on hand and robot data with a common vision encoder and policy network. Despite distinct action spaces for human and robot, our model enforces a shared representation to enable performance scaling with human embodiment data, outperforming existing methods that treat human and robot data separately.

We empirically evaluate \method on three challenging long-horizon manipulation tasks in the real world: continuous object-in-bowl, clothes folding, and grocery packing (Fig.~\ref{fig:tasks}). Our results demonstrate that \method significantly enhances task performance across all scenarios, with relative improvements of up to 200\%. Notably, we observe that \method exhibits generalization to objects and scenes encountered exclusively in human data.  Finally, we analyze the scaling properties of \method, and found learning from an additional hour of hand data significantly outperforms training from an additional hour of robot data.

%% file: sections/relatedWorks.tex
\section{Related Works}\label{relatedWorks}
\noindent\textbf{Imitation Learning:}
Imitation Learning (IL) has been used to perform diverse and contact-rich manipulation tasks \cite{NIPS2013_e53a0a29, finn2017oneshotvisualimitationlearning, robomimic2021}. 
Recent advancements in IL have led to the development of pixel-to-action IL models, which directly map raw visual inputs to low-level robot control \cite{zhao2023act, chi2024diffusionpolicy}. These visual IL models have demonstrated impressive reactive policies \cite{young2020visualimitationeasy, wang2023mimicplay}. Scaling these models has displayed strong generalization in works such as RT1 and RT2 \cite{brohan2023rt1, brohan2023rt2}. However, these methods remain labor and resource-intensive, for instance RT1 required 17 months of data collection and 13 robots \cite{brohan2023rt1}. 
Our work proposes a learning framework that takes advantage of scalable human embodiment demonstrations, which has the potential to be larger and more diverse than any dataset consisting of robot demonstrations alone.

\noindent\textbf{Learning from Video Demonstrations:}
To satisfy the data requirements of pixel to action IL algorithms, many recent works leverage human data because it is highly scalable. Human data is used at different levels of abstraction, where some works use human videos from internet-scale datasets to pretrain visual representations \cite{nair2022r3m, radosavovic2023real, ma2022vip}. Other works use human videos to more explicitly understand scene dynamics through point track prediction, intermediate state hallucination in pixel space, or affordance prediction \cite{xiong2021learningwatchingphysicalimitation, wen2024anypointtrajectorymodelingpolicy, bharadhwaj2023hopman, bharadhwaj2024track2act, bahl2023affordanceshumanvideosversatile}.  And finally, recent works use hand trajectory prediction as a proxy for predicting robot actions~\cite{wang2023mimicplay}.  {While these approaches leverage hand data, they often have separate modules to process hand and robot data.  Instead, by fully leveraging the rich information provided by \aria including on-board SLAM, our method is able to unify and treat human and robot data as equals and co-train from both data sources with a single end-to-end policy.}

\noindent\textbf{Data Collection Systems:}
Various methods have been used to scale robot data. Low-cost devices such as the Space Mouse offer sensitive and fine-grained teleoperation of robotic manipulators \cite{mandlekar2020learning, robomimic2021, Dhat2024Using3M, chi2024diffusionpolicy, zhu2023viola}. Further works improve intuitive control through virtual reality systems such as the VR headset \cite{arunachalam2023holo, george2023openvrteleoperationmanipulation, 9013428, cheng2024open, he2024omnih2o}. Recent systems like ALOHA and GELLO increase ergonomics for low-cost and fine-grained bimanual manipulation tasks through a leader-follower teleoperation interface \cite{zhao2023act, wu2024gellogenerallowcostintuitive} or exoskeletons~\cite{fang2023low, yang2024ace}.
Other works attempt to collect human embodiment data with rich information like 3D action tracking, but existing systems face tradeoffs. Those which leverage rich information are either not portable (e.g., static camera~\cite{sivakumar2022robotic, wang2023mimicplay, jain2024vid2robot, fu2024humanplus}) or ergonomic (e.g., require a hand-held gripper~\cite{chi2024universalmanipulationinterfaceinthewild, shafiullah2023bringing} or body-worn camera~\cite{wang2024dexcapscalableportablemocap, papagiannis2024r+}), which prevent the passive scalability of the data collection system. 
Along these lines, our approach captures egocentric video and 3D hand tracking data, but via the ergonomic form factor of Project Aria Glasses~\cite{engel2023projectarianewtool}. 
While other wearable data collection methods like VR headsets capture hand positions to teleoperate a robot, our system does not require a robot at all.
This system has the potential to passively scale~\cite{grauman2024ego}, as adoption of similar consumer-grade devices continue to rise.

\noindent\textbf{Cross-embodiment Policy Learning:} 
Advances in cross-embodiment learning show that large models trained on datasets with diverse robot embodiments are more generalizable \cite{openx2024}. Some approaches aim to bridge the embodiment gap through observation reprojection \cite{chen2024miragecrossembodimentzeroshotpolicy},  action abstractions \cite{huang2020policycontrolallshared}, and policies conditioned on embodiment, \cite{yang2024pushinglimitscrossembodimentlearning}. Recent works view cross-embodiment learning as a domain adaptation problem \cite{yang2023polybottrainingpolicyrobots}. Our work argues that human data should be treated as another embodiment in transfer learning.

%% file: sections/method.tex
\input{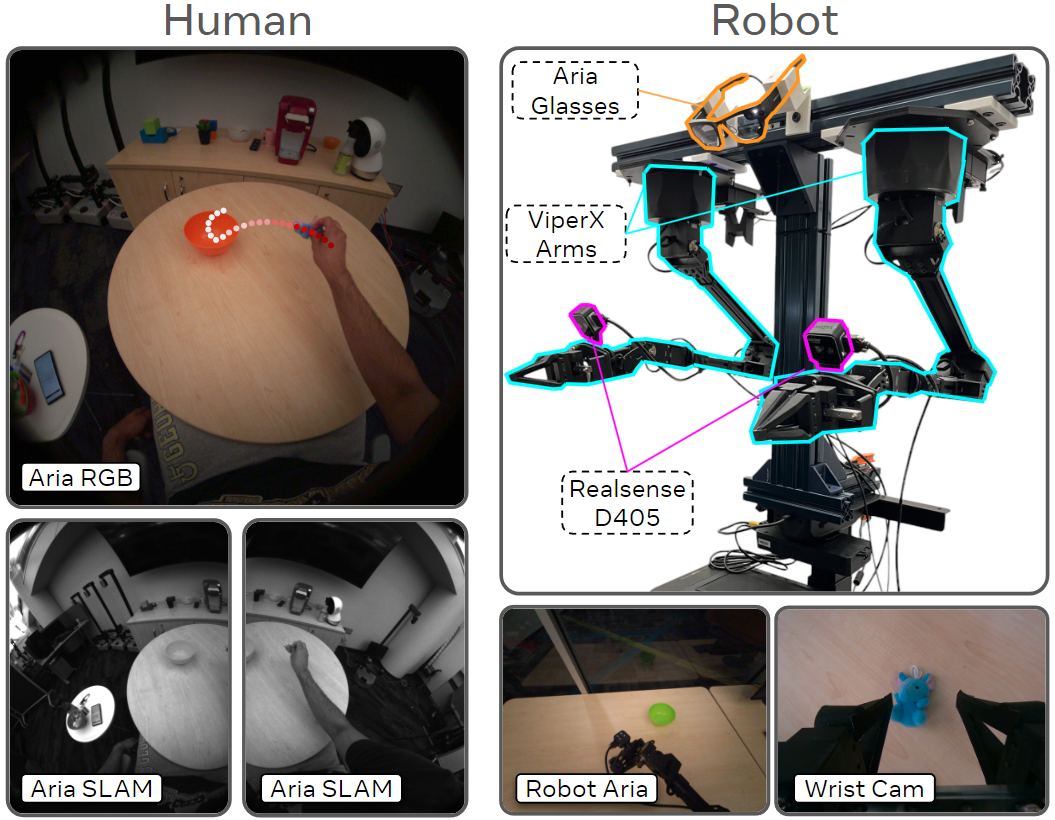}
\section{\method}\label{method}

\method is a full-stack framework that captures and learns from both egocentric human embodiment data and robot data. We detail each component of our pipeline below, starting with our hardware setup for human and robot data collection (Sec.~\ref{ssec:hardware}), followed by our methods for processing and aligning the data from both sources (Sec.~\ref{ssec:data_process}), and finally our unified policy architecture (Sec.~\ref{ssec:policy}). Our design choices throughout are motivated by making human embodiment data as suitable for robot learning as tele-operated robot data is.

\subsection{Data Collection Systems and Hardware Design}
\label{ssec:hardware}
\noindent \textbf{\aria for egocentric demonstration collection.}
An ideal system for human data needs to capture rich information about the scene, while remaining passively scalable.  Such a system should be wearable, ergonomic, capture a wide FOV, track hand positions, device pose, and more.

\method fills this gap by building on top of the Project \aria~\cite{engel2023projectarianewtool}. \aria are head-worn devices for capturing multimodal egocentric data. The device assumes an ergonomic glasses form factor that weighs only 75g, permitting long wearing time and passive data collection. Our work leverages the front-facing wide-FoV RGB camera for visual observation and two mono-color scene cameras for device pose and hand tracking (See Fig.~\ref{fig:hardware} for sample data). In particular, the side-facing scene cameras track hand poses even when they move out of the main RGB camera’s view, significantly mitigating the challenges posed by humans' natural tendency to move their head and gaze ahead of their hands during sequential manipulation tasks.

Further, there are large scale data collection efforts underway with Project Aria~\cite{grauman2022ego4dworld3000hours, ma2024nymeriamassivecollectionmultimodal}, and the devices are made available broadly to the academic community through an active research partnership program.  In the future, our system can enable users to seamlessly merge data they collect with these large datasets. %
Ultimately, we present a system that enables passive yet feature-rich human data collection to help scale up robot manipulation.

\noindent\textbf{Low-cost bimanual manipulator.}
To effectively utilize egocentric human embodiment data, a robot manipulator should be capable of moving in ways that resemble human arm movements. Prior works often rely on table-mounted manipulators such as the Franka Emika Panda~\cite{frankaemikapanda}. While these systems are capable, they differ significantly from human arms in terms of kinematics. Moreover, their substantial weight and inertia necessitate slow, cautious movements due to safety concerns, largely preventing them from performing manipulation tasks at speeds comparable to humans.
In response to these limitations, we have purpose-built a bimanual manipulator that is lightweight, agile, and cost-effective. Drawing inspiration from the ALOHA system~\cite{zhao2023act}, our robot setup comprises two 6-DoF ViperX 300 S arms with Intel Realsense D405 wrist cameras, mounted in an inverted configuration on a height-adjustable rig as the torso (Fig ~\ref{fig:hardware}), kinematically mimicking the upper body of a human. The ViperX arms are lean and relatively similar in size to human arms, contributing to their enhanced agility.
The entire rig can be assembled for less than \price excluding the ViperX arms (the BOM will be made available). We also built a leader robot rig to collect teleoperation data, similar to ALOHA~\cite{zhao2023act}.

Further, as our method jointly learns visual policies from egocentric human and robot data, it is essential to align the visual observation space. Thus in addition to alignment through data post-processing (Sec.~\ref{ssec:data_process}), we directly match the camera hardware by using a second pair of \aria as the main sensor for the robot, which we have mounted directly to the top of the torso at a location similar to that of human eyes (Fig.~\ref{fig:hardware}).  This enables us to mitigate the observation domain gap associated with the camera devices, including FOVs, exposure levels, and dynamic ranges.

\subsection{Data Processing and Domain Alignment}
\label{ssec:data_process}
To train unified policies from both human and robot data, \method bridges three key human-robot gaps: (1) unifying action coordinate frames, (2) aligning action distributions, and (3) mitigating visual appearance gaps.

\textbf{Raw data streams.} 
We stream raw sensor data from the hardware setup as described in Sec.~\ref{ssec:hardware}. \aria worn by the human and robot generate ego-centric RGB image streams. In addition, the robot generates two wrist camera streams. For proprioception, we leverage the Aria Machine Perception Service (MPS)~\cite{projectaria_mps_summary} to estimate 3D poses of both hands $\prescript{H}{}{p} \in \mathbb{SE}(3) \times \mathbb{SE}(3)$. Robot proprioception data includes both its end effector poses $\prescript{R}{}{p} \in \mathbb{SE}(3) \times \mathbb{SE}(3)$ and joint positions $\prescript{R}{}{q}\in \mathbb{R}^{2\times7}$ (including the gripper jaw joint position). We in addition collect joint-space actions $\prescript{R}{}{a}^q \in \mathbb{R}^{2\times7}$ for teleoperated robot data.  

\input{tables/data_streams}

\textbf{Unifying human-robot data coordinate frames.} Robot action and proprioception data typically use fixed reference frames (e.g., camera or robot base frame). However, egocentric hand data from moving cameras breaks this assumption. To unify the reference frames for joint policy learning, we transform both human hand and robot end effector trajectories into camera-centered stable reference frames.
Following the idea of predicting action chunks~\cite{chi2024diffusionpolicy,zhao2023act}, we aim to construct action chunks $a^p_{t:t+h}$ for both human hand and robot end effector. To simplify the notation, we describe the single-arm case that generalizes to both arms. The raw trajectory is a sequence of 3D poses $[p^{F_t}_t, p^{F{t+1}}_{t+1}, ... p^{F{t+h}}_{t+h}]$, where $F_i$ denotes the coordinate frame of the camera when estimating $p_i$. $F_i$ remains fixed for the robot but changes constantly for human egocentric data. Our goal is to construct $a^p_{t:t+h}$ by transforming each position in the trajectory into the observation camera frame $F_{t}$. This allows the policy to predict actions without considering future camera movements.  For human data, we use the MPS visual-inertial SLAM to obtain the \aria pose $T_{F_{i}}^W \in \mathbb{SE}(3)$ in the world frame and transform the action trajectory:
$$\prescript{H}{}{a}^p_i = [(T_{F_t}^W)^{-1} T_{F_i}^W p^{F_i}_i \quad \text{for} \quad i \in [t, t+1, ..., t+h]]$$
A sample trajectory is visualized in Fig.~\ref{fig:hardware} (top-left). Robot data is transformed similarly using the fixed camera frame estimated by hand-eye calibration. By creating a unified reference frame, we enable the policy to learn from action supervisions regardless of whether they originate from human videos or teleoperated demonstrations. %

\input{figures/dists}

\textbf{Aligning human-robot pose distributions.} Despite aligning hand and robot data via hardware design and data processing, we still observe differences in the distributions of hand and robot end effector poses in the demonstrations collected. These discrepancies arise from biomechanical differences, task execution variations, and measurement precision disparities between human and robotic systems. Without mitigating this gap, the policy tends to learn separate representations for the two data sources~\cite{yang2024pushing, hejna2024re}, preventing performance scaling with human data. To address this, we apply Gaussian normalization individually to end effector (hand) poses and actions from each data source, as shown in Fig.~\ref{fig:dists}. %
Echoing~\cite{hejna2024re}, we found this simple technique to be empirically effective (Sec.~\ref{sec:results}), though we plan to explore alternatives such as action quantization~\cite{brohan2023rt1} in the future.

\input{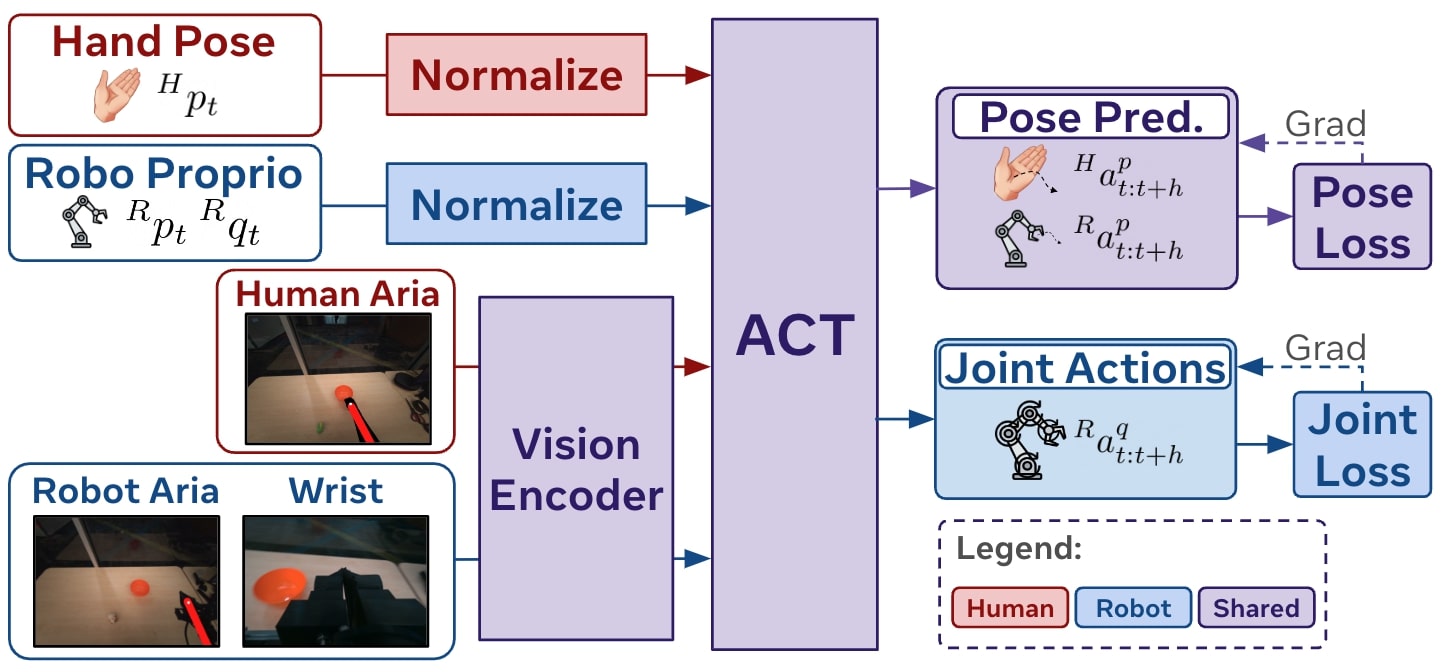}
\textbf{Bridging visual appearance gaps.} Despite aligning sensor hardware for capturing robot and human data, there still exists a large visual appearance gap between human hands and robots. Previous works have acknowledged this gap and attempt to occlude or remove the manipulator in visual observation \cite{zhou2021manipulatorindependentrepresentationsvisualimitation}, \cite{bahl2022humantorobotimitationwild}. %
{We follow similar ideas and mask out both the hand and the robot via SAM~\cite{ravi2024sam2segmentimages}} and overlay a red line to indicate end-effector directions (Fig.~\ref{fig:dists}).  The SAM point prompts are generated by the robot end effector and human hand poses transformed to image frames.

\input{figures/tasks}
\subsection{Training Human-Robot Joint Policies}
\begin{algorithm}
\caption{Joint Human-Robot Policy Learning}
\label{alg:joint_policy_learning}
\begin{algorithmic}[1]
\Require Human dataset $\mathcal{D}_H$, Robot dataset $\mathcal{D}_R$
\State Initialize shared transformer encoder $f_{enc}(\cdot)$, pose decoder $f^p(f_\text{enc}(\cdot))$, and joint decoder $f^q(f_\text{enc}(\cdot))$
\For{iteration $n = 1, 2, \ldots$}
    \State // Human data
    \State Sample $(I_t, p_t, a^p_{t:t+h})$ from $\mathcal{D}_H$
    \State Predict $\hat{a}^p_{t:t+h}$ from $f_p(f_\text{enc}(I_t, p_t))$
    \State $\mathcal{L}^H_p = \text{MSE}(\hat{a}^p_{t:t+h}, a^p_{t:t+h})$
    \State // Robot data
    \State Sample $(I_t, p_t, q_t, a^p_{t:t+h}, a^q_{t:t+h})$ from $\mathcal{D}_R$
    \State Predict $\hat{a}^q_{t:t+h}$ from $f_q(f_\text{enc}(I_t, p_t, q_t))$
    \State Predict $\hat{a}^p_{t:t+h}$ from $f_p(f_\text{enc}(I_t, p_t, q_t))$
    \State $\mathcal{L}^R_q = \text{MSE}(\hat{a}^q_{t:t+h}, a^q_{t:t+h})$
    \State $\mathcal{L}^R_p = \text{MSE}(\hat{a}^p_{t:t+h}, a^p_{t:t+h})$
    \State // Joint policy update
    
    \State Update $f_{enc}, f^p, f^q$ with $\mathcal{L}^H_p + \mathcal{L}^R_p+ \mathcal{L}^R_q$
\EndFor
\end{algorithmic}
\end{algorithm}

\label{ssec:policy}
Existing approaches often opt for hierarchical architectures, where a high-level policy trained on human data conditions a low-level policy outputting robot actions~\cite{wang2023mimicplay,bharadhwaj2023hopman}. However, this approach is inherently limited by the performance of the low-level policy, which does not directly benefit from large-scale human data.
To address this limitation, we propose a simple architecture (illustrated in Fig.~\ref{fig:arch}) that learns from unified data and promotes shared representation.  
Our model builds upon ACT~\cite{zhao2023act}, but the design is general and can be applied to other transformer based imitation learning algorithms.

A critical challenge in this unified approach is the choice of the robot action space. While the robot end-effector poses are more semantically similar to human hand pose than robot-joint positions, it is difficult to control our robot with end-effector poses via a cartesian-based controller (e.g., differential IK) because the 6 DoF ViperX arms offer low solution redundancy. Empirically, we found that robots often encounter singularities or non-smooth solutions in a trajectory. Consequently, we opt for joint-space control (i.e., use predicted joint action  $\hat{a}^q_{t:t+h}$ to control the robot), while leveraging pose-space prediction to learn joint human-robot representation. Note that the need both pose- and joint-space predictions is specific to our robot hardware, and more capable robots that better support end-effector space control can eliminate the need for predicting joint-space actions.

Specifically, all parameters in the policy are shared besides the two shallow input and output heads. The input heads transform the visual and proprioceptive embeddings before passing to the policy transformer.  The policy transformer processes these features, and the two output heads transform the transformer's latent output into either pose or joint space predictions.  The pose loss supervises both human and robot data via $^Ha^p$ and $^Ra^p$, whereas the joint action loss only supervises robot data $^Ra^q$. Since the two branches are separated by only one linear layer, we effectively force the model to learn joint 
representations for both domains. The algorithm is summarized in Alg.~\ref{alg:joint_policy_learning}. Table~\ref{tab:data_streams} summarizes the data used for training.

%% file: figures/hardware.tex
\begin{figure*}[t]
\centering
\includegraphics[width=\textwidth]{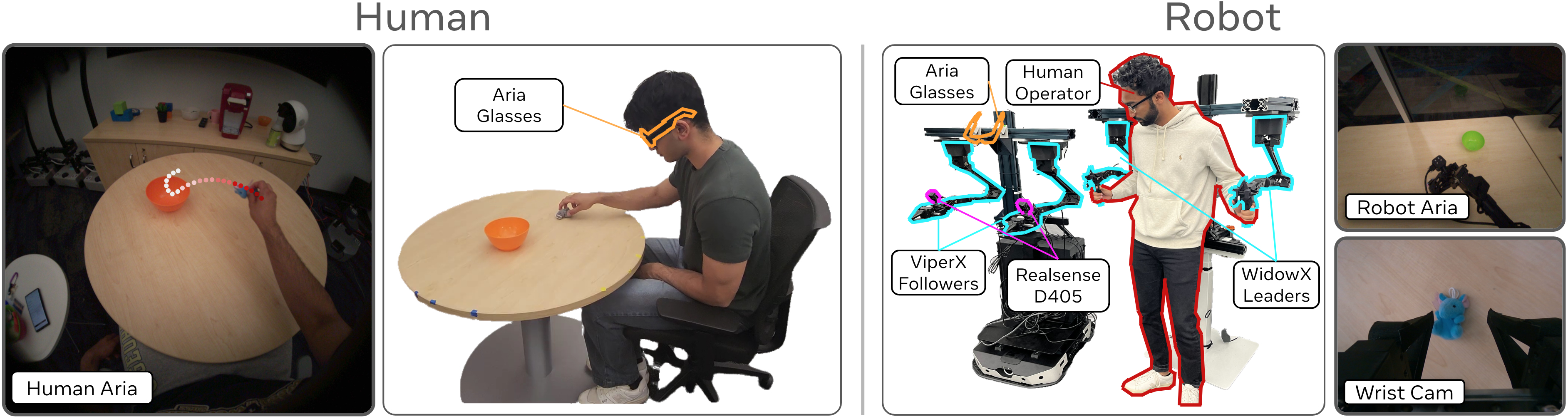}
\caption{Our human data system uses \aria to capture Egocentric RGB and uses its side SLAM cameras to localize the device and track hands. The robot consists of two Viper X follower arms with Intel RealSense D405 wrist cameras, controlled by two WidowX leader arms. Our robot uses identical Aria glasses as the main vision sensor to help minimize the camera to camera gap.}
\label{fig:hardware}
\vspace{-10pt}
\end{figure*}

%% file: tables/data_streams.tex
\begin{table}[t]
    \centering
\begin{tabular}{lll}
    \toprule
    \textbf{Source} & \textbf{Type} & \textbf{Data} \\
    \midrule
    \textbf{Human $\mathcal{D}_H$} & Image & Egocentric view \\
                                   & Proprio & 3D hand poses ($^Hp$) \\
                                   & Action & Normalized hand tracks ($^Ha^p$) \\
    \midrule
    \textbf{Robot $\mathcal{D}_R$} & Image & Egocentric + wrist views \\
                                   & Proprio & EEF poses ($^Rp$), Joint positions ($^Rq$) \\
                                   & Action & EEF actions ($^Ra^p$), Joint actions ($^Ra^q$) \\
    \bottomrule
\end{tabular}
    \caption{Comparison of human and robot data streams}
    \label{tab:data_streams}
\end{table}

%% file: figures/dists.tex
\begin{figure}[t]
\centering
\includegraphics[width=\linewidth]{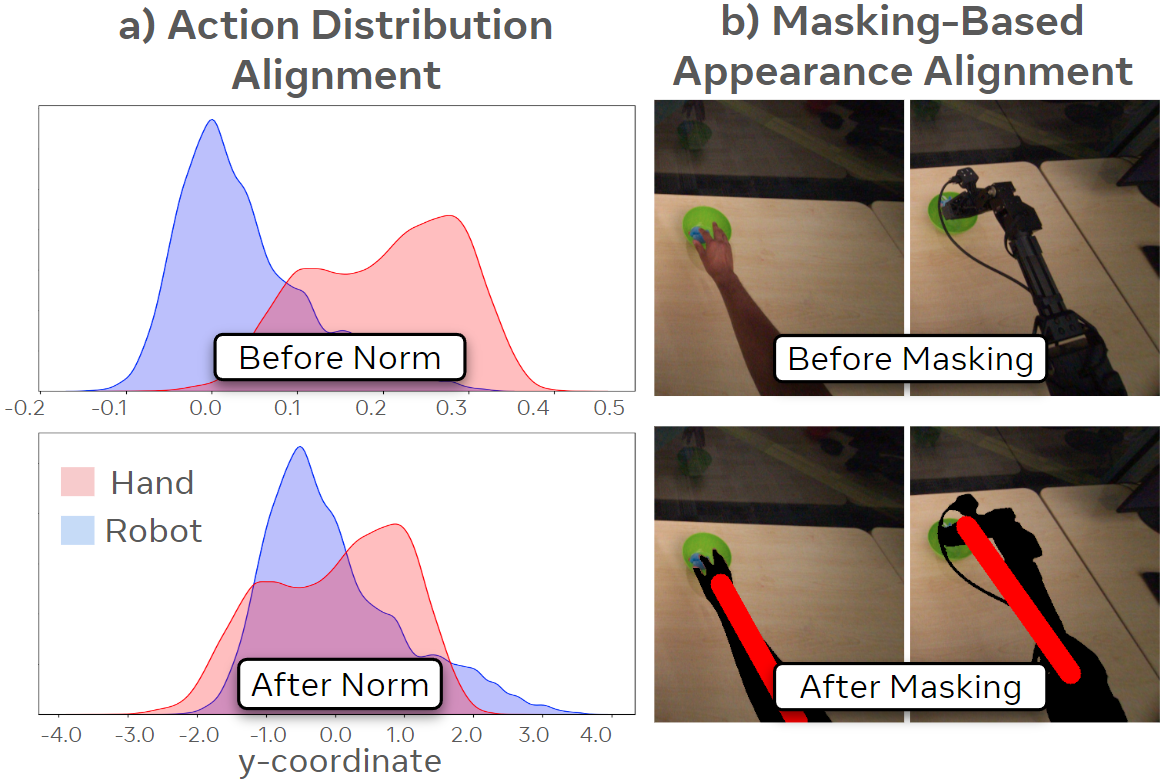}
\caption{\textbf{a) Action normalization:} The pose distributions are different between hand and robot data, specifically in the $y$ (left-right) dimension.  We apply Gaussian normalization individually to the hand and robot pose data before feeding them to the model.  \textbf{b) Visual masking:} To help bridge the appearance gap of human and and the robot arm, we apply a black mask to the hand and robot via SAM, then overlay a red line onto the image.
}

\label{fig:dists}
\end{figure}

%% file: figures/arch.tex
\begin{figure}[t]
\centering
\includegraphics[width=\linewidth]{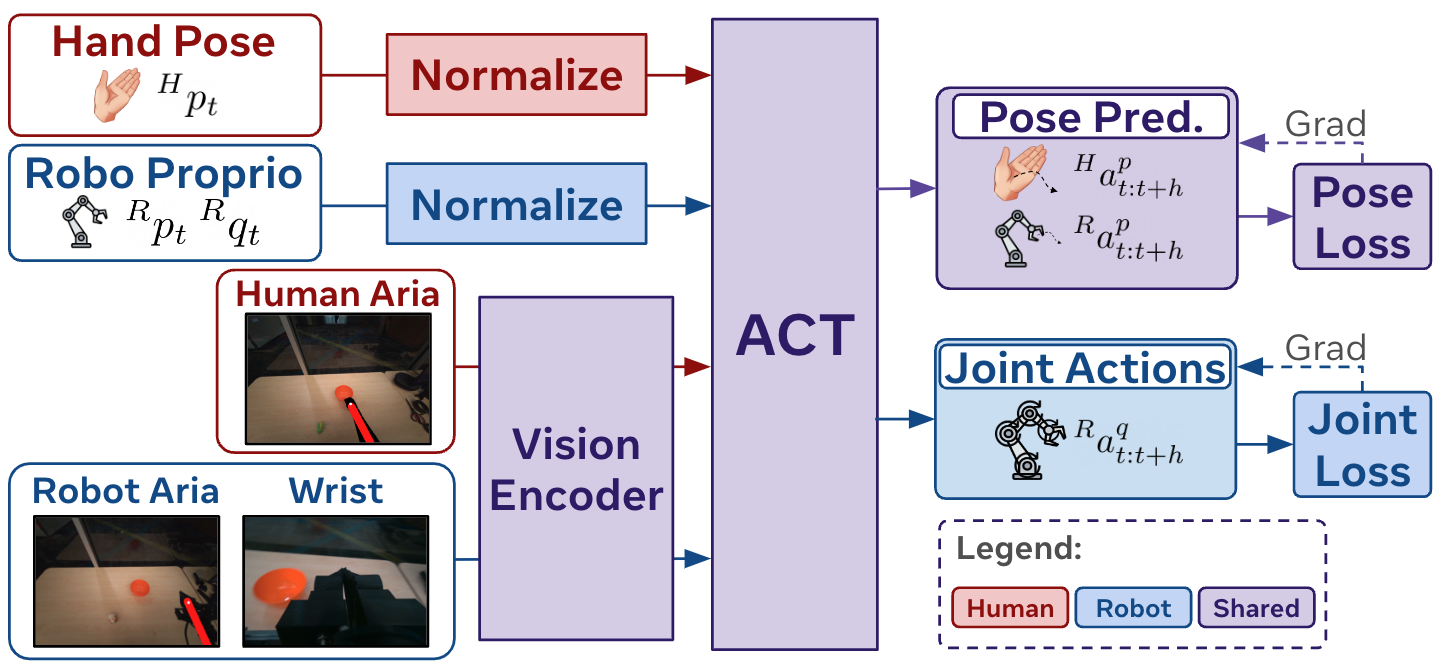}
\caption{
Architecture of the joint human-robot policy learning framework. The model processes normalized hand and robot data through shared vision and ACT encoders, outputting pose predictions for both human and robot data, and joint actions for robot data. The framework uses masked images to mitigate human-robot appearance gaps and incorporates wrist camera views for the robot. %
}
\label{fig:arch}
\end{figure}

%% file: figures/tasks.tex
\begin{figure*}[t]
\centering
\includegraphics[width=\textwidth]{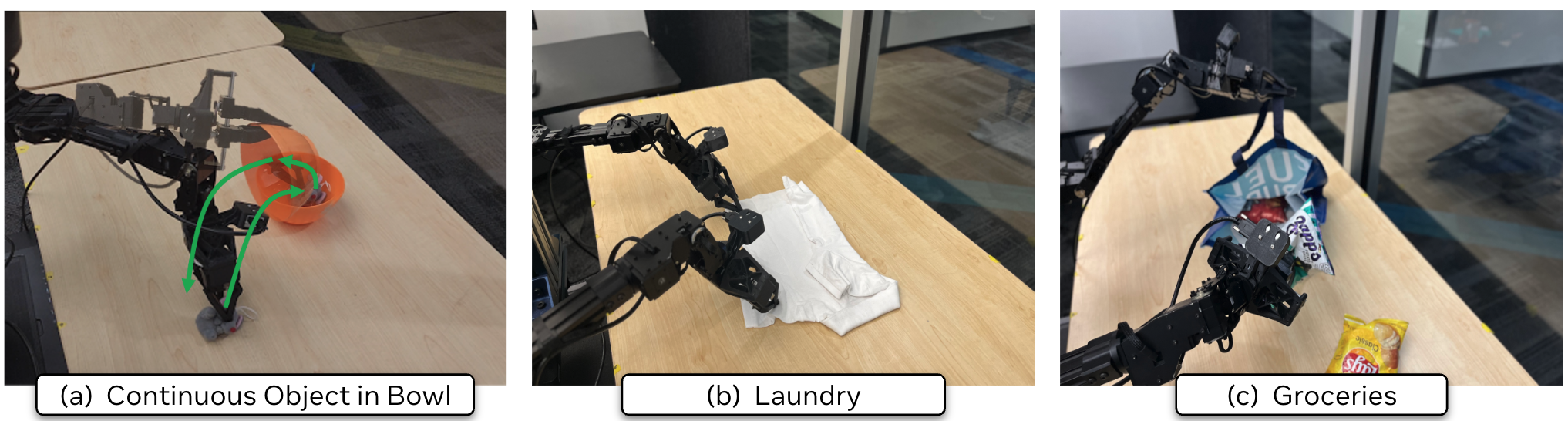}
\caption{We evaluate \method across three real world, long-horizon manipulation tasks. See Sec.~\ref{sec:experimentalSetup} for description. %
}
\label{fig:tasks}
\vspace{-10pt}
\end{figure*}

%% file: sections/experimentalSetup.tex
\section{Experiments}

We aim to validate three key hypotheses. \textbf{H1}: \method is able to leverage human embodiment data to boost in-domain performance for complex manipulation tasks. \textbf{H2}: Human data helps \method generalize to new objects and scenes. \textbf{H3}: {Given sufficient initial robot data, it is more valuable to collect additional human data than additional robot data.}

\subsection{Experiment Setup}
\label{sec:experimentalSetup}
\noindent\textbf{Tasks.}
We select a set of long-horizon real world tasks to evaluate our claims.  Our tasks require precise alignment, complex motions, and bimanual coordination (Fig.~\ref{fig:tasks}).

\input{tables/data}

\textit{Continuous Object-in-Bowl}: The robot picks a small plush toy (about 6cm long), places it in a bowl, picks up the bowl to dump the object onto the table, and repeats continuously for 40 seconds. We randomly choose from a set of 3 bowls and 5 toys which randomly positioned on the table within a 45cm x 60cm range. The task stress-tests precise manipulation, spatial generalization, and robustness in long-horizon execution.  We award \textbf{Pts} each time the toy is placed in a bowl, or the bowl is emptied.  We perform 45 total evaluation rollouts across 9 bowl-toy-position combinations.  %

\textit{Laundry}: A bimanual task that requires the robot to fold a t-shirt placed with random pose in a {90cm $\times$ 60cm} range and a rotation range of $\pm 30\deg$. The robot must use both arms to fold the right side sleeve, the left side sleeve, then the whole shirt in half.  We award \textbf{Pts} for each of these stages, and calculate Success Rate \textbf{(SR)} based as the percentage of runs where all stages were successful.  We perform 40 total evaluation rollouts across 8 shirt-position combinations.

\textit{Groceries}: The robot fills a grocery bag with 3 packs of chips.  It uses its left arm to grab the top side of the bag handle to create an opening, then uses the right arm to pick the chip packs and places them into the bag. The task requires high-precision manipulation (picking up a deformable bag handle) and robustness in long-horizon rollout. We award \textbf{Pts} for picking the handle and for each pack placed in the grocery bag.  We report \textbf{SR} as the percentage of runs where all three packs were successfully placed in the bag, and \textbf{Open Bag} as the percentage of runs where the handle of the bag was grasped, which is a difficult stage of this task.  We perform 50 evaluations across 10 bag positions.

\simar{We detail the amount of data collected for each task in Table~\ref{tab:data}.  While collecting robot data in particular, we make sure to randomly perturb the robot's position, which we found empirically to improve robustness.  For human data, we note that while tasks like \textit{Continuous Object-in-Bowl} were particularly easy to scale, tasks like \textit{Groceries} were slower because of resetting time.}

\noindent\textbf{Baselines.}
To evaluate that \method can improve in-domain success rate by leveraging human data, we benchmark against ACT~\cite{zhao2023act}, a state of the art imitation learning algorithm.  Further, we compare against Mimicplay~\cite{wang2023mimicplay}, a recent state of the art method that learns planners from human data to guide low-level policies, to show that our unified architecture learns more effectively from human and robot data. For fair comparisons, we implement Mimicplay with the same Transformer backbone as our method, \simar{and we removed goal conditioning because \method is designed for single-task policies. Since \method contains architectural changes to ACT, namely the simultaneous joint and pose action prediction, we also benchmark against \method (0\% Human).  This helps us conclude that improvements come from leveraging human data rather than the architecture.}

%% file: tables/data.tex
\begin{table}[t]
\centering
\caption{Data collection overview for both Human(H) and Robot(R) data.  We report both the number(\#) of total task demonstrations and the time(min) took to collect them.}
\vspace{0.1in}
\begin{tabular}{lcccccc}
    \toprule
    \bf Task & \bf H & \bf H & \bf H& \bf R & \bf R & \bf R  \\
    \bf  & \bf \# & \bf min & \bf \#/min & \bf \# & \bf min & \bf \#/min  \\
    \toprule
    Object-in-Bowl & 1400 & 60 & 23 & 270 & 120& 2\\
    Groceries & 160 & 80 & 2 & 300 & 300& 1\\
    Laundry & 590 & 100 & 6 & 430 & 300& 1\\
    \bottomrule
    \end{tabular}
\label{tab:data}
\vspace{-10pt}
\end{table}

%% file: tables/success_rate.tex
\begin{table}[t]
\centering
\caption{Quantitative results for 3 real-world tasks. We report task success rates (\%) and performance scores (pts) for all tasks and bag grabbing rate for the Groceries tasks. 
}
\vspace{0.1in}
\setlength{\tabcolsep}{1.1mm}{
\begin{tabular}{lc|cc|ccc}
    \toprule
    \bf Method & \multicolumn{1}{c}{\bf Bowl} & \multicolumn{2}{c}{\bf Laundry} & \multicolumn{3}{c}{\bf Groceries}\\
    \cmidrule(lr){2-2} \cmidrule(lr){3-4} \cmidrule(lr){5-7}
    & \bf Pts & \bf Pts & \bf SR & \bf Pts & \bf SR & \bf Open Bag \\
    \toprule
    ACT~\cite{zhao2023act} & 39 & 82 & 55\% & 82 & 22\% & 54\%\\
    Mimicplay~\cite{wang2023mimicplay} & 71 & 78 & 50\% & 53 & 8\% & 40\% \\
    \method (w/o human) & 68 & 104 & 73\% & 92 & 28\% & 60\%\\
    \method & \bf128 & \bf114 & \bf88\% & \bf110 & \bf30\% & \bf70\%\\
    \bottomrule
\end{tabular}
}
\label{tab:successrate}
\vspace{-10pt}
\end{table}

%% file: tables/ablations.tex
\begin{table}[t]
\centering
\caption{\textbf{Ablations} - We ablate our method and report final task performance on the Object-in-Bowl task. 
}
\vspace{0.1in}
\begin{tabular}{lc}
    \toprule
    \bf Method & \bf Cotrained (Points)\\
    \toprule
    \method & 128\\
    \method w/o Line & 112\\
    \method w/o Line and Mask & 95\\
    \method w/o Action Norm & 79\\
    \method w/o Hand Data & 68\\
    \bottomrule
    \end{tabular}
\label{tab:ablations}
\end{table}

%% file: sections/results.tex
\subsection{Results}\label{sec:results}

\input{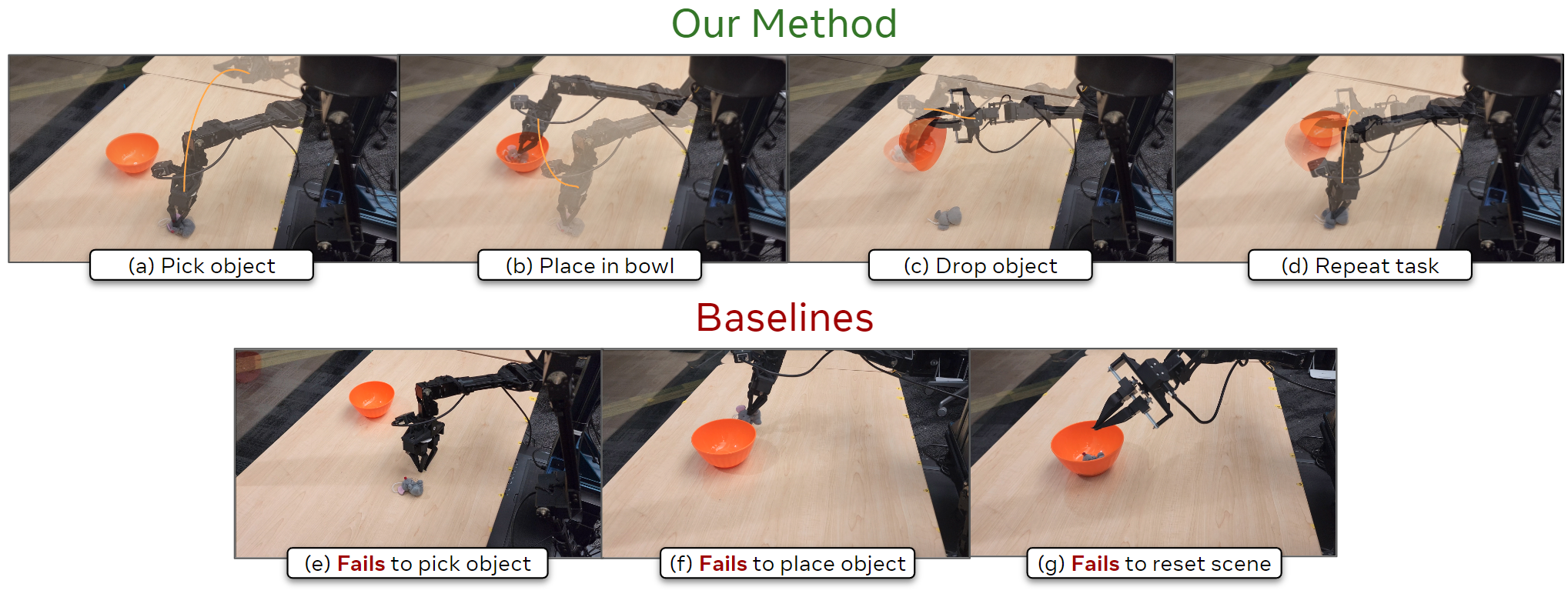}

\noindent \textbf{\method improves in-domain task performance}.  Across all tasks we observed a relative improvement in score of 34-228\%, and an improvement in absolute task success rate from 8-33\% over ACT.  Our largest improvement is on the \emph{Cont. Object-in-Bowl} task, in which we yield a 228\% improvement in task score over ACT.  We observe the baselines often miss the toy or bowl by a few inches, which seems to indicate that our use of hand data helps the policy precisely reach the toy. We show qualitative results in Fig.~\ref{fig:qualSuccess}.

To ensure this increase was due to leveraging hand data rather than architectural changes, we compare to \method (0\% human).  We observe a 10-88\% improvement in score and 2-15\% improvement in success rate.

\input{figures/generalization}

\noindent \textbf{\method enables generalization to new objects and even scenes.}  We evaluate our method on two domain shifts: attempting to fold shirts of an unseen color, and performing the \emph{Cont. Object-in-Bowl} task in an entirely different scene. As shown in Fig.~\ref{fig:generalization}, we observe that ACT struggles on shirts of unseen colors (25\% SR) whereas \method fully retains its performance (85\% SR).  %
Surprisingly, by learning from human data in a new scene (unseen background and lighting), \method is able to generalize to this new environment without \textit{any} additional robot data, scoring 63 points. In contrast, Mimicplay, which had access to the same information but instead leverages a hierarchical framework for using hand data only scored 4 points.  This suggests that our architecture promotes joint hand-robot representation, whereas hierarchical architectures pose a generalization bottleneck.

\noindent\textbf{Scaling human vs. robot data.} 
To investigate the scaling effect of human and robot data sources on performance, we conducted additional data collection for the \emph{Cont. Object-in-bowl} task. As illustrated in Fig. 7, EgoMimic trained on 2 hours of robot data and 1 hour of human data significantly outperforms ACT trained on 3 hours of robot data (128 vs 74 points). Notably, one hour of human data yields 1400 demonstrations, compared to only 135 demonstrations from an hour of robot data. These results demonstrate EgoMimic's ability to effectively leverage the efficiency of human embodiment data collection, leading to a more pronounced scaling effect that substantially boosts task performance beyond what is achievable with robot data alone.  We note that \method at 2 hours of robot data outperforms ACT at 2 hours of robot data, so some improvement is attributed to architecture.

\input{figures/scaling}

\noindent\textbf{Ablation studies.}
We ablate our approach to demonstrate the importance of each design decision on the \textit{Object-in-Bowl} task (Table~\ref{tab:ablations}).  \simar{First, removing action normalization results in a 38\% drop in task score.  This highlights the importance of action distribution alignment for co-training.  Next, we ablate away the visual techniques, specifically masking out the hand and robot, as well as drawing the red overlay on the image.  Removing these components resulted in 13 and 26\% drops respectively.  Finally, \method trained without any hand data, yields a large 47\% drop, which highlights how effective hand-robot co-training is on our stack.
}

%% file: figures/qualSuccess.tex
\begin{figure}[t]
\centering
\includegraphics[width=\columnwidth]{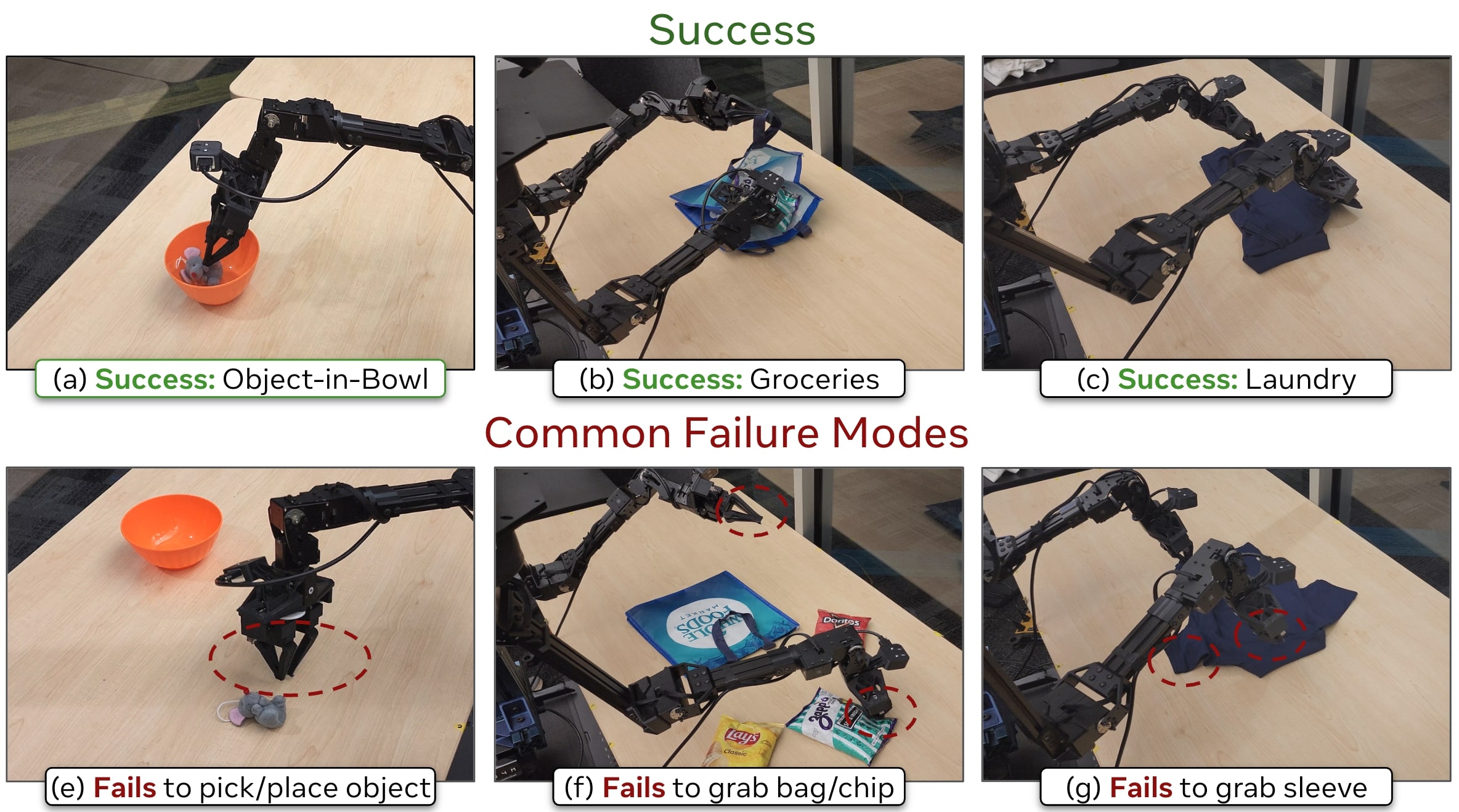}
\caption{We highlight \method's success, as well as failure modes, for instance \textbf{(e)} failure to correctly align with the toy, \textbf{(f)} failure to grasp the bag's handle, or \textbf{(g)} policy only grabs 1 side of the shirt.
\textbf{\method} reduces the frequency of these failure modes, improving success rates by 8-33\% over the baselines.}
\label{fig:qualSuccess}
\vspace{-10pt}
\end{figure}

%% file: figures/generalization.tex
\begin{figure}[t]
\centering
\includegraphics[width=0.5\textwidth]{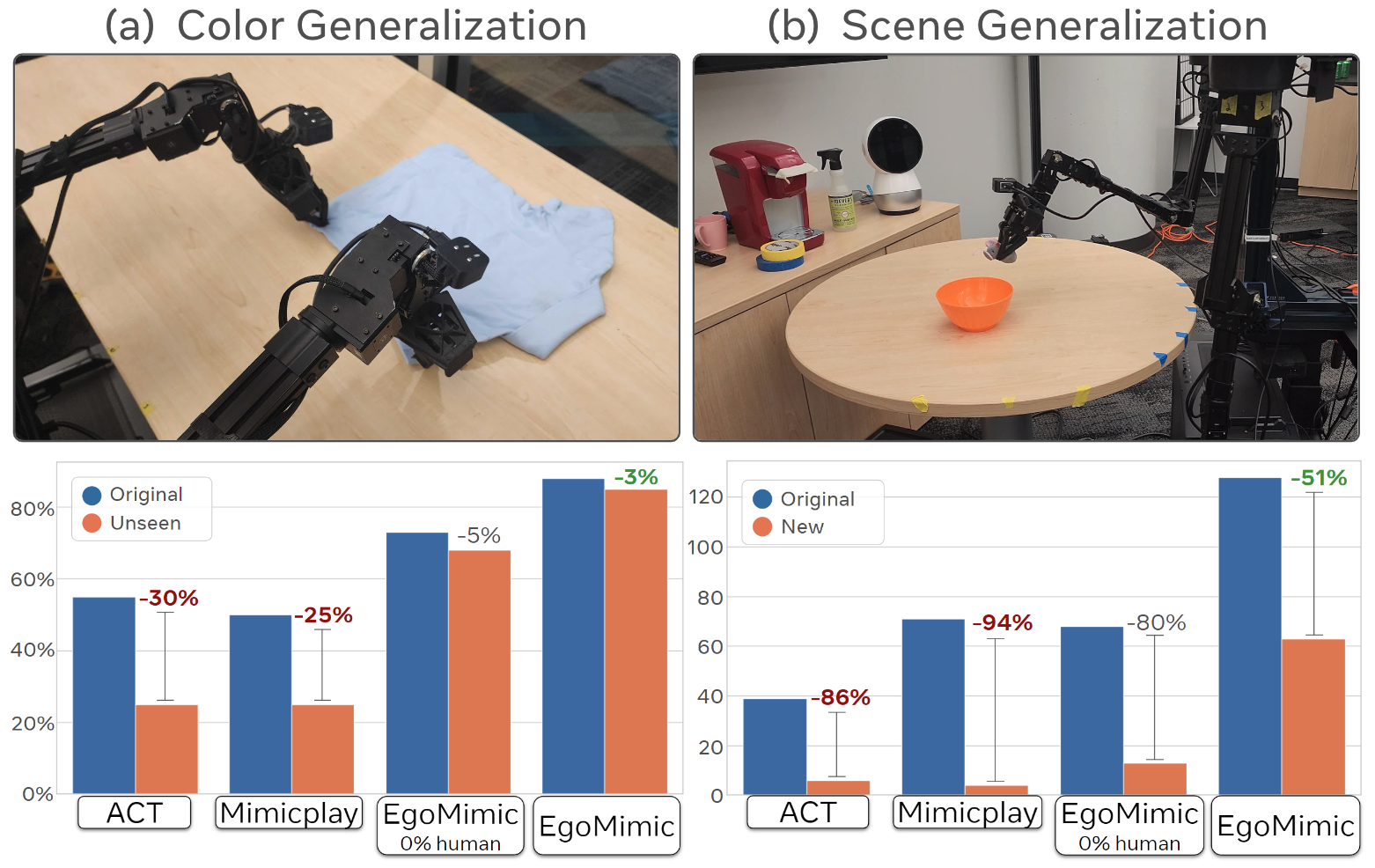}
\caption{\textbf{Evaluation Results on Policy Generalization.} (a) We evaluate the policy on the laundry task using unseen cloth colors and report the success rate for each method. (b) We test the policy on the Object-in-Bowl task in unseen scenes.}
\label{fig:generalization}
\end{figure}

%% file: figures/scaling.tex
\begin{figure}[t]
  \centering
  \includegraphics[width=\linewidth]{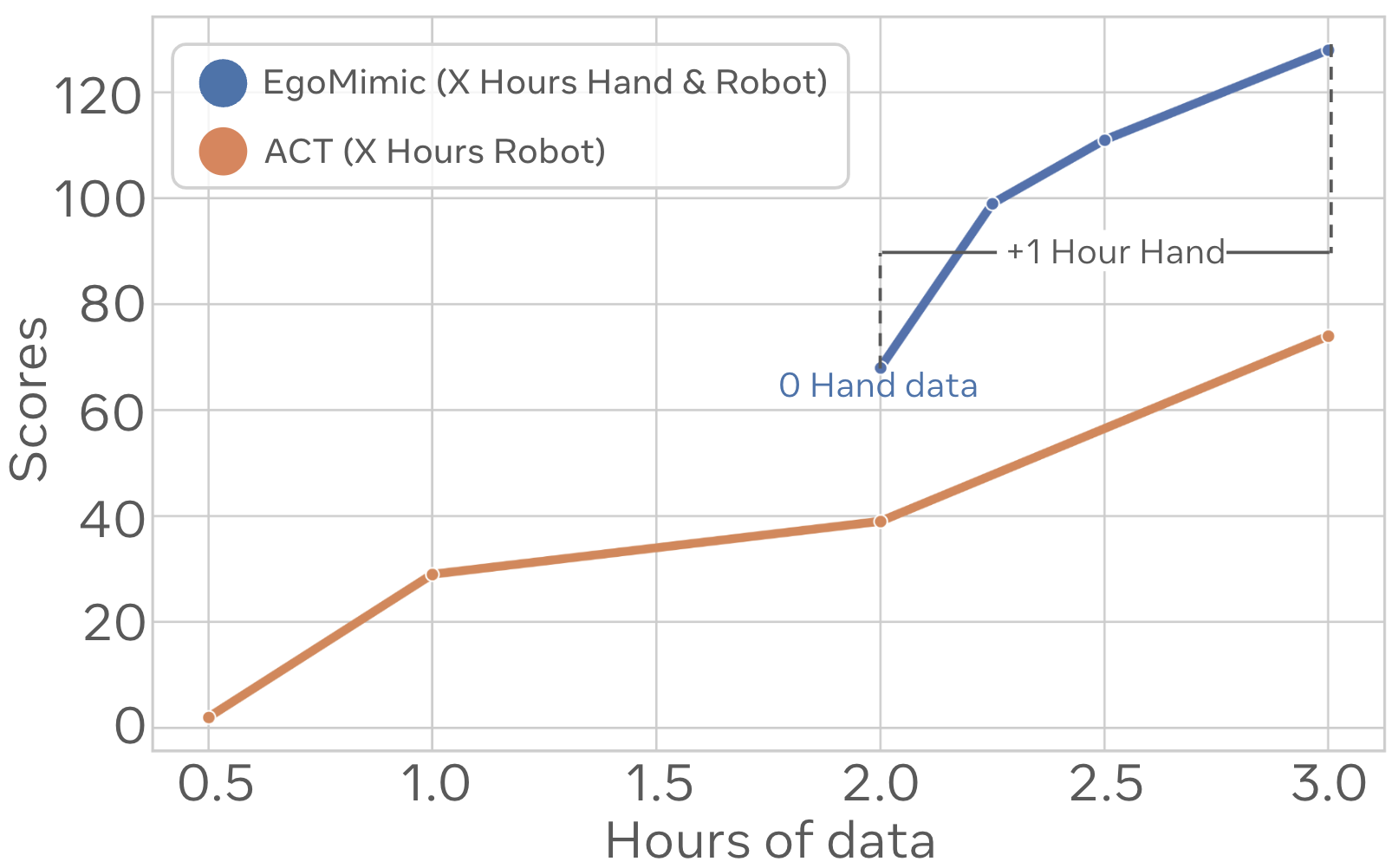}
  \caption{\textbf{Scaling robot vs. human data. }\method trained on 2 hours robot data + 1 hour hand data (Blue) strongly outperforms ACT~\cite{zhao2023act} trained on 3 hours of robot data (Orange).}
\label{fig:scaling}
\vspace{-10pt}
\end{figure}

%% file: sections/conclusion.tex
\section{Conclusions}

\simar{
We presented \method, a framework to co-train manipulation policies from human egocentric videos and teleoperated robot data. By leveraging Project Aria glasses, a low-cost bimanual robot setup, cross-domain alignment techniques, and a unified policy learning architecture, \method improves over state-of-the-art baselines on three challenging real-world tasks and shows generalization to new scenes as well as favorable scaling properties.  %
For future work, we plan to explore the possibility of generalizing to new robot embodiments and entirely new behaviors demonstrated only in human data, such as folding pants instead of shirts. Overall, we believe our work opens up exciting new venues of research on scaling robot data via passive data collection.
}

%% file: sections/appendix.tex
\input{figures/handvsRobot}
\section{Appendix}
\subsection{Data Processing and Domain Alignment}
Humans and teleoperated robots complete tasks at different speeds. To enable joint training of human and robot data, we must align these two sources of data temporally.  Following Mimicplay~\cite{wang2023mimicplay}, we ``slow down'' the human data, and we found empirically that a factor of 4 sufficiently aligned both domains.  
Specifically, for robot data we construct joint and pose based actions over a four second horizon 
but for human data we use a 1 second horizon.
For both domains, our action chunk size is 100, meaning we construct 100 future actions spaced evenly over the horizon.
This alignment is independent of data recording frequencies, where human data is recorded at 30hz and robot data is recorded at 50 hz.

To co-train on both human and robot data, we individually normalize the proprioception and actions for both embodiments (as shown in Fig.~\ref{fig:dists}).  Given proprioception $p_t\in\mathbb{R}^d$ where $d$ depends on embodiment, we normalize by subtracting the dataset mean and dividing by standard deviation $$norm(p_t) = (p_t - \mu_{p})/\sigma_{p}.$$  We perform the identical calculation to normalize actions $a_{t:t+h}\in\mathbb{R}^{d\times 100}$.

\input{figures/detailedArch}
To bridge the appearance gap between human hand and robot arm, we visually mask each embodiment via SAM2~\cite{ravi2024sam2segmentimages}, and overlay a red line on these masks to enhance alignment (Fig.~\ref{fig:dists}).
For the robot, we first use forward kinematics to compute the 3D coordinates of key joints in robot frame $$p_t^R = FK(q_t) \in \mathbb{R}^{3\times 3},$$ including the wrist, gripper and the forearm. These 3D coordinates are then projected onto the image frame via camera intrinsics ($I_{cam}^{pixels}$) and extrinsics ($T_R^{cam}$) to obtain 2D keypoints in pixel space $$p_t^{pixel} = I_{cam}^{pixels} T_R^{cam} p_t^R\in \mathbb{R}^{3\times 2},$$ 
which are used to prompt SAM2~\cite{ravi2024sam2segmentimages} to generate a mask of the robot arm. After obtaining the mask, we draw a red line on the masked area from the gripper to the elbow in the RGB image. 
For the human data, a similar process is followed, where SAM2 is prompted using the 3D coordinates of the human hand to generate a mask. A red line is then drawn along the hand’s contour, from the bottom right to top left corner of the contour's bounding box. 

During training, both the robot arm and human hand are masked to align their visual representations. During evaluation, SAM2 is run in real time on a desktop to mask the robot arm and apply the same red line overlay. This approach enables better visual alignment between the robot and human hand, facilitating more effective model generalization across human and robotic tasks.

\subsection{Aria Machine Perception Services (MPS)}
We leveraged MPS to process human data from the Aria glasses.  The raw data from Aria contains timestamped sensor information from the glasses, namely RGB camera, SLAM cameras, IMU, eye tracking cameras, microphone, and more. The raw data is uploaded to the MPS server, where the cloud-hosted service estimates device pose via SLAM, a semi-dense pointcloud of the environment, hand tracking relative to the device frame, and even eye gaze.  The MPS returns SLAM as a timestamped CSV of device poses in world frame and hand tracking as a timestamped CSV of cartesian positions in the time-aligned device frame.  These hand positions are each in a distinct reference frame due to head movements, so we project future actions to the current device coordinate frame (described in Sec.~\ref{ssec:data_process}).
We use the undistorted Aria RGB camera data paired with the hand tracking and SLAM information to construct an hdf5 file compatible for training in robomimic~\cite{robomimic2021}. 

\subsection{Training Human-Robot Joint Policies}
We depict our algorithm in detail in Fig.~\ref{fig:detailedArch}.  At each step we sample a batch of hand data as well as a batch of robot data, and pass each through our unified architecture.  \method performs Z-score normalization to hand and robot proprioception and actions individually.  The normalized proprioception is passed through a linear layer to produce a proprioception token.  Alongside the proprioception, the top down views from hand and robot are passed through a SAM based masking module.  These images, along with the robot wrist views are passed through a shared Resnet18 visual encoder which produces visual tokens.  Finally, we add an additional style token $z$ from our CVAE encoder which is not depicted, but directly follows ACT~\cite{zhao2023act}.  All these tokens, are passed through a transformer encoder decoder architecture.  The transformer decoder's hidden output is passed through a linear decoder depending on the output type, producing pose actions $\hat{a}^p$ or joint based actions $\hat{a}^j$.

For batches of robot data, we calculate $$L_{robot}=L_1(^R\hat{a}^p, ^Ra^p) + L_1(^R\hat{a}^j, ^Ra^j) + KL$$ and for hand data we have $$L_{hand}=L_1(^H\hat{a}^p, ^Ha^p) + KL$$ where $KL$ is the CVAE latent regularizer as in ACT~\cite{zhao2023act}. This yields $L = L_{robot} + L_{hand}$ which we optimize at each step.

We leverage the transformer's flexible input sequence to account for differences in the number of visual observations based on the modality; specifically we have wrist images in robot data but not hand data.  When the wrist images are present, we concatenate additional tokens to our transformers input sequence as in ACT~\cite{zhao2023act}.  In our experiments, we found that this strategy was sufficient to effectively co-train on both hand and robot data, although we plan to experiment with more sophisticated cross-embodiment learning techniques like HPTs~\cite{wang2024hpt}.

We note that the human data lacks information for the grasping action, since Aria only records hand pose.  Thus, the grasping action is supervised only via the robot joint prediction loss $L_1(^R\hat{a}^j, ^Ra^j)$, where the gripper is represented as another joint.

\subsection{Training Details}
We list the hyperparameters for \method in Table~\ref{tab:act_params}.  All models were trained for 120000 iterations with global batch size of 128 across 4 A40 gpus, which takes about 24 hours.  Our code is implemented in the robomimic framework~\cite{robomimic2021}. More details in Table~\ref{tab:act_params}
\input{tables/act_hyperparams}
\input{figures/rollouts}
\subsection{Mimicplay Implementation}
For our implementation of MimicPlay~\cite{wang2023mimicplay}, we closely follow the original setup, training the high-level planner and low-level control policy separately.

First, we train a ResNet-18 based high-level encoder using a Gaussian Mixture Model (GMM) to generate 3D trajectories, as described in the original work. The high-level encoder is trained on both human and robot data to predict 3D trajectories.

Once the high-level encoder is trained, we extract the latent representation from the ResNet-18 encoder (i.e., the high-level planner) and use it as the style variable $z$, which is passed to the transformer encoder-decoder Fig.~\ref{fig:detailedArch}.  The low-level ACT policy is then trained solely on robot data with this additional input from the high level policy as guidance.
\input{tables/mimicplay_hyperparams}

\subsection{Policy Rollout}
\input{tables/freqs}
We rollout our policy with inference at 1hz and control at 25hz on a desktop with a an NVIDIA RTX 4090 GPU.  The predicted action horizon is 4 seconds, with the first second of predicted actions executed in receding-horizon style.  
All the robot's sensors update at 50hz with the exception of the Aria camera which streams at 30fps.

%% file: figures/handvsRobot.tex
\twocolumn[{%
  \renewcommand\twocolumn[1][]{#1}%
  \vspace{-0.5cm}
  \includegraphics[width=\textwidth]{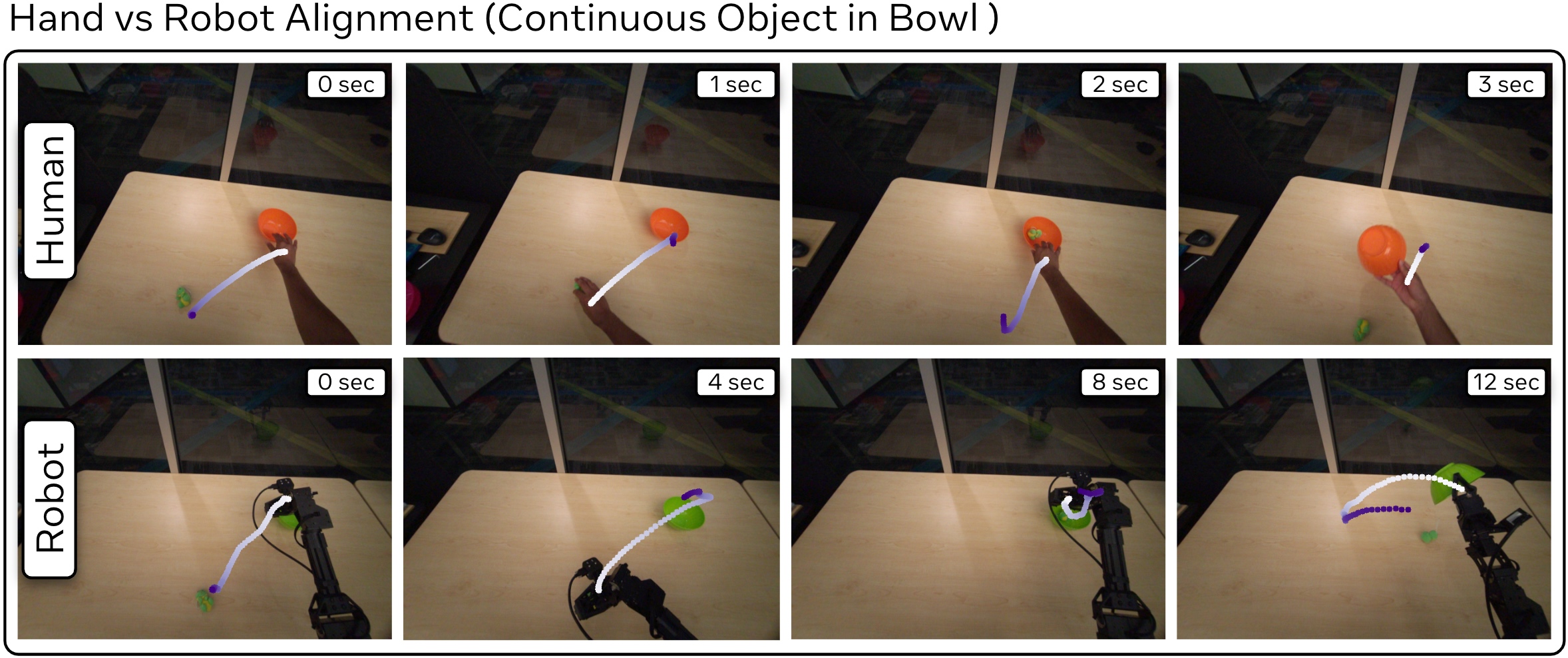}
  \vspace{-0.45cm}
  \captionof{figure}{Here we visualize hand and robot data from out dataset side by side with ground truth actions overlayed (purple).  Note that the actions are of similar length, despite the hand traveling much faster than the robot.}
  \label{fig:handvsrobot}
  \vspace{0.5cm}
}]

%% file: figures/detailedArch.tex
\begin{figure*}[t]
\centering
\includegraphics[width=\linewidth]{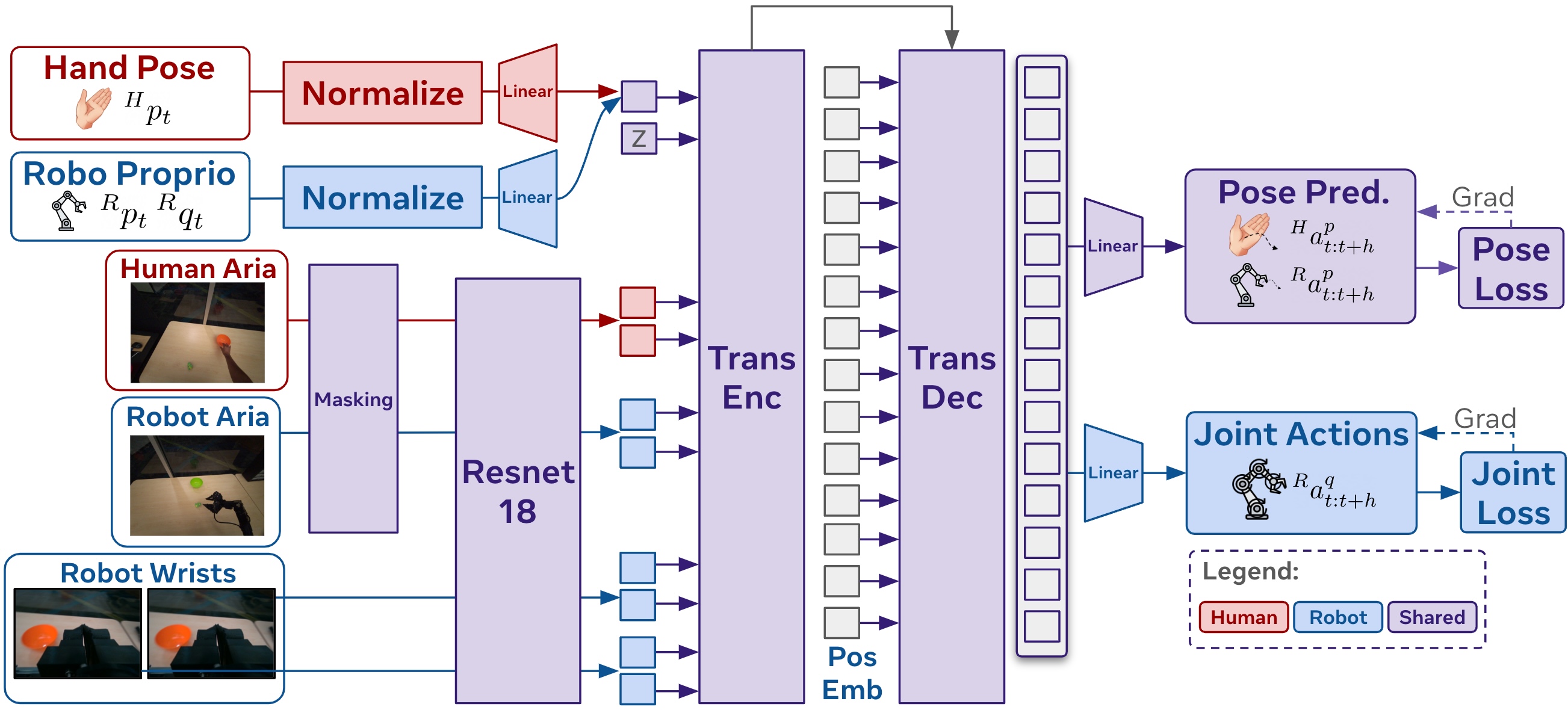}
\caption{Detailed Architecture of \method.}
\vspace{-10pt}
\label{fig:detailedArch}
\end{figure*}

%% file: tables/act_hyperparams.tex
\begin{table}[t]
\centering
\caption{Training details - EgoMimic}
\vspace{0.1in}
\begin{tabular}{lc}
    \toprule
    \bf Policy & ACT \\
    \bf Batch Size & 128 \\
    \bf Optimizer & adamw \\
    \bf Learning rate (initial) &  5e-5 \\
    \bf Decay factor & 1 \\
    \bf Scheduler & Linear \\
    \bf Encoder layers & 4 \\
    \bf Decoder layers & 7 \\
    \bf Hidden dim & 512 \\
    \bf Feedforward dim & 3200 \\
    \bf No. of heads & 8 \\
    \bf Data Augmentations & Color Jitter\\
    \bottomrule
\end{tabular}
\label{tab:act_params}
\end{table}

%% file: figures/rollouts.tex
\begin{figure*}[t]
\centering
\includegraphics[width=\textwidth]{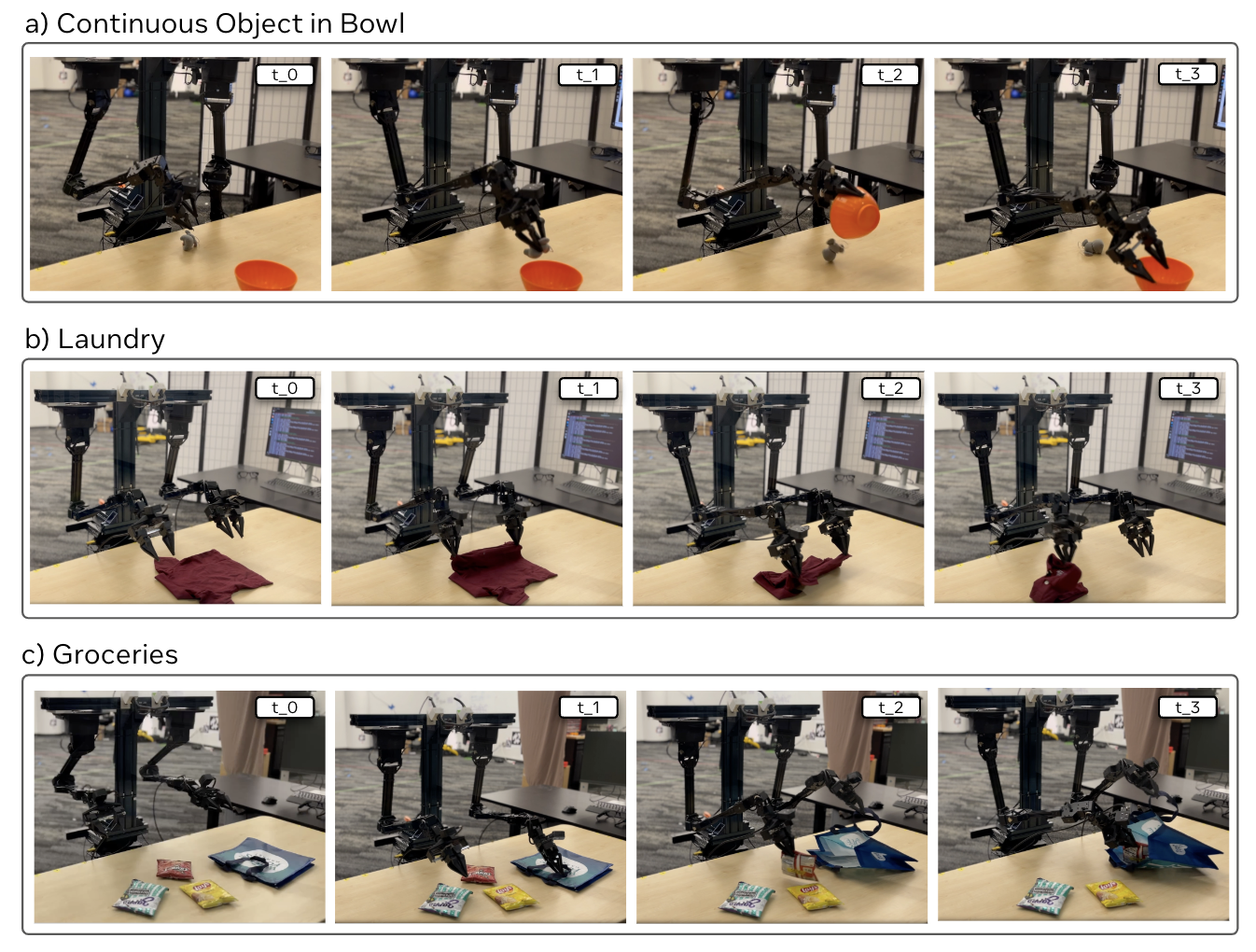}
\caption{Qualitative successes of \method on each of our three tasks.}
\label{fig:rollouts}
\vspace{-10pt}
\end{figure*}

%% file: tables/mimicplay_hyperparams.tex
\begin{table}[t]
\centering
\caption{Training details - Mimicplay}
\vspace{0.1in}
\begin{tabular}{lc}
    \toprule
    \bf High-level & Resnet18 \\
    \ \ \ Learning rate (initial) & 0.0001 \\
    \ \ \ Decay factor & 0.1 \\
    \ \ \ Batch size & 50 \\
    \ \ \ GMM modes & 5 \\
    \bf Low-level & ACT \\
    \ \ \ Learning rate (initial) &  5e-5 \\
    \ \ \ Optimizer & adamw \\
    \ \ \ Decay factor & 1 \\
    \ \ \ Scheduler & Linear \\
    
    \bottomrule
\end{tabular}
\label{tab:mimicplay_params}
\end{table}

%% file: tables/freqs.tex
\begin{table}[t]
\centering
\caption{Data recording and rollout rates for Human and Robot data.  We ``slow down'' human data by 0.25 to account for differences in task execution speeds.}
\vspace{0.1in}
\begin{tabular}{lcc}
    \toprule
    \bf Type & \bf Human (Hz) & \bf Robot (Hz) \\
    \midrule
    \bf Recording & 30 & 50 \\
    \midrule
    \bf Rollout (Inference) & - & 1 \\
    \bf Rollout (Control) & - & 25 \\
    \bottomrule
\end{tabular}
\label{tab:recording_rollout}
\end{table}